\documentclass[11pt,a4paper]{article}
\usepackage[hyperref]{acl2021}
\usepackage{times}
\usepackage{latexsym}
\usepackage{multirow}
\usepackage{amsmath}
\usepackage{amssymb}
\usepackage{float}
\usepackage{pgfplots}
\usepackage{pgfplotstable}
\pgfplotsset{compat=1.7}
\usepackage{tikz}
\usepackage{graphicx}
\usepackage{graphics}
\usepackage{booktabs}
\usepackage{subcaption}
\usepackage{flushend}
\usepackage{soul}
\usepackage{color}
\usepackage{microtype}

\newcommand{\hlc}[2][yellow]{{%
    \colorlet{foo}{#1}%
    \sethlcolor{foo}\hl{#2}}%
}

\newcommand\dunderline[3][-1pt]{{%
  \sbox0{#3}%
  \ooalign{\copy0\cr\rule[\dimexpr#1-#2\relax]{\wd0}{#2}}}}

\aclfinalcopy 

\title{
Measuring and Increasing Context Usage in\\ 
Context-Aware Machine Translation
}

\author{
Patrick Fernandes$^{1,2,3}$ \qquad
Kayo Yin$^{1}$ \qquad
Graham Neubig$^{1}$ \qquad
André F. T. Martins$^{2,3,4}$\\
$^1$Language Technologies Institute, Carnegie Mellon University, Pittsburgh, PA \\
$^2$Instituto Superior Técnico \& LUMLIS (Lisbon ELLIS Unit), Lisbon, Portugal \\
$^3$Instituto de Telecomunicações, Lisbon, Portugal \\
$^4$Unbabel, Lisbon, Portugal \\
 {\small \texttt{\{pfernand, kayoy, gneubig\}@cs.cmu.edu} \quad \texttt{andre.t.martins@tecnico.ulisboa.pt}}
}
\date{}

\begin{document}
\maketitle

\begin{abstract}

Recent work in neural machine translation has demonstrated both the necessity and feasibility of using \emph{inter-sentential} context --- context from sentences other than those currently being translated. However, while many current methods present model architectures that theoretically \emph{can} use this extra context, it is often not clear how much they \emph{do} actually utilize it at translation time.
In this paper, we introduce a new metric, \emph{conditional cross-mutual information}, to quantify the usage of context by these models. Using this metric, we measure how much document-level machine translation systems use particular varieties of context.
We find that target context is referenced more than source context, and that conditioning on a longer context has a diminishing effect on results. We then introduce a new, simple training method, \emph{context-aware word dropout}, to increase the usage of context by context-aware models. Experiments show that our method increases context usage and that this reflects on the translation quality according to metrics such as BLEU and COMET, as well as performance on anaphoric pronoun resolution and lexical cohesion contrastive datasets.\footnote{ \href{https://github.com/neulab/contextual-mt}{https://github.com/neulab/contextual-mt}}

\end{abstract}

\section{Introduction}

While neural machine translation (NMT) is reported to have achieved human parity in some domains and language pairs \citep{DBLP:journals/corr/abs-1803-05567}, these claims seem overly optimistic and no longer hold with document-level evaluation \citep{toral-etal-2018-attaining, laubli-etal-2018-machine}.
Recent work on \textit{context-aware} NMT attempts to alleviate this discrepancy by incorporating the surrounding context sentences (in either or both the source and target sides) in the translation system.
This can be done by, for example, feeding context sentences to standard NMT models \cite{tiedemann-scherrer-2017-neural}, using different encoders for context \cite{zhang-etal-2018-improving}, having cache-based memories \cite{tu-etal-2018-learning}, or using models with hierarchical attention mechanisms \citep{miculicich-etal-2018-document,maruf-etal-2019-selective} --- more details in \S\ref{sec:context-aware}.
While such works report gains in translation quality compared to sentence-level baselines trained on small datasets, recent work has shown that, in more realistic high-resourced scenarios, these systems fail to outperform simpler baselines with respect to overall translation accuracy, pronoun translation, or lexical cohesion \cite{lopes-etal-2020-document}.

\begin{figure}
  \includegraphics[width=1\columnwidth]{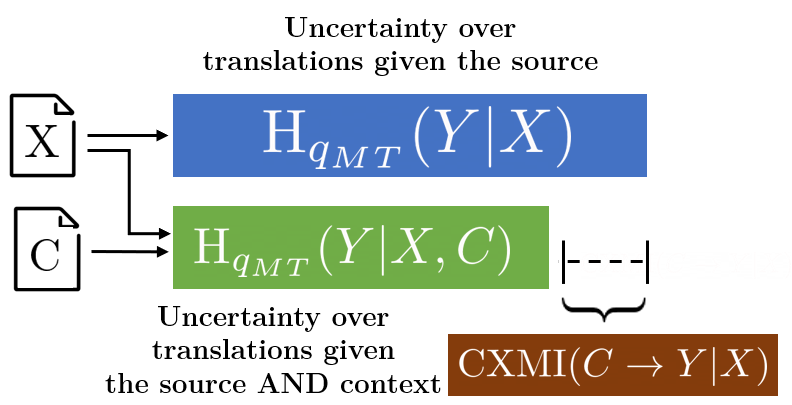}
    \caption{Illustration of how we can measure context usage by a model $q_{MT}$ as the amount of information gained when a model is given the context $C$ and source $X$ vs when the model is only given the $X$.}
    \label{fig:cxmi-illustration}
    \vspace{-1.5em}
\end{figure}

We hypothesize that one major reason for these lacklustre results is due to the fact that models with the architectural capacity to model cross-sentential context do not necessarily learn to do so when trained with existing training paradigms.
However, even quantifying model usage of context is an ongoing challenge; while contrastive evaluation has been proposed to measure performance on inter-sentential discourse phenomena \cite{muller-etal-2018-large, bawden-etal-2018-evaluating}, this approach is confined to a narrow set of phenomena, such as pronoun translation and lexical cohesion. A toolbox to measure the impact of context in broader settings is still missing.

To address the limitations above, we take inspiration from the recent work of \citet{bugliarello-etal-2020-easier} and propose a new metric, \textbf{conditional cross-mutual information} (\textsc{CXMI}, \S\ref{sec:xmi}), to measure quantitatively how much context-aware models actually use the provided context by comparing the model distributions over a dataset with and without context. Figure \ref{fig:cxmi-illustration} illustrates how it measures context usage. This metric applies to \emph{any} probabilistic context-aware machine translation model, not only the ones used in this paper. We release a software package to encourage the use of this metric in future context-aware machine translation research.
We then perform a rigorous empirical analysis of the \textsc{CXMI} between the context and target for different context sizes, and between source and target context. We find that: (1) context-aware models use some information from the context, but the amount of information used does not increase uniformly with the context size, and can even lead to a \emph{reduction} in context usage; (2) target context seems to be used more by models than source context.

\begin{table}
\vspace{\baselineskip}
\centering \small
\setlength{\tabcolsep}{0.5ex}
\begin{tabular}{p{0.25\columnwidth}p{0.71\columnwidth}}
\toprule
\textbf{Source:} &  \textit{The Church is merciful…} \\
& It always welcomes the misguided lamb. \vspace{0.2em} \\
\midrule
\textbf{Target:}  & \textit{Die Kirche ist barmherzig…} \vspace{0.2em}\\
\centering\textit{Baseline} & \textcolor{red}{Es} heisst die fehlgeleiteten Schäflein immer willkommen. \\ 
\centering \textit{Context-Aware} & \textcolor{red}{Es} heisst die fehlgeleiteten Schäflein immer willkommen.  \\
\centering { \textit{Context-Aware  w/ our method} } & \textcolor{blue}{Sie} heisst die fehlgeleiteten Schäflein immer willkommen. \\ 
\bottomrule
\end{tabular}
\caption{Example where context (italic) is needed to correctly translate the pronoun ``it''. Both the sentence-level baseline and context-aware model fail to correctly translate it while the context-aware model trained with \textit{\textsc{CoWord} dropout} correctly captures the context.\label{tab:examples} }
\label{fig:example}
\end{table}

Given the findings, we next consider how to encourage models to use more context.
Specifically, we introduce a simple but effective variation of word dropout \cite{sennrich-etal-2016-edinburgh} for context-aware machine translation, dubbed  \textbf{\textsc{CoWord} dropout} (\S\ref{sec:source_dropout}). Put simply, we randomly drop words from the \textit{current} source sentence by replacing them with a placeholder token. Intuitively, this encourages the model to use extra-sentential information to compensate for the missing information in the current source sentence. 
We show that models trained with \textsc{CoWord} dropout not only increase context usage compared to models trained without it but also improve the quality of translation, both according to standard evaluation metrics (BLEU and COMET) and according to contrastive evaluation based on inter-sentential discourse phenomena such as anaphoric pronoun resolution and lexical cohesion (\S\ref{sec:experiments}, Table \ref{fig:example}).



\section{Context-Aware Neural Machine Translation}
\label{sec:context-aware}

We are interested in learning a system that translates documents consisting of multiple sentences between two languages.%
\footnote{Here, a ``document'' could be an actual document but it could also represent other contextual collections of text, such as a sequence of dialogue utterances.}
More formally, given a corpus of parallel documents in two languages, $\mathcal{D} = \{D_1, ..., D_N\}$, where each document is a sequence of source and target sentences, $D = \{(x^{(1)}, y^{(1)}),...,(x^{(K)}, y^{(K)})\}$, we are interested in learning the mapping between the two languages.

We consider the typical (auto-regressive) neural machine translation system $q_\theta$ parameterized by $\theta$. The probability of translating $x^{(i)}$ into $y^{(i)}$ given the context of the sentence $C^{(i)}$ is 
$$
q_\theta(y^{(i)} | x^{(i)}, C^{(i)}) = \prod_{t=1}^T q_\theta(y^{(i)}_t | x^{(i)}, y^{(i)}_{<t}, C^{(i)})
$$
where $y^{(i)}_t$ represents the $t\textsuperscript{th}$ token of sentence $y^{(i)}$.
This context can take various forms.
On one end, we have the case where no context is passed, $C^{(i)}=\varnothing$, and the problem is reduced to sentence-level translation. On the other end, we have the case where all the source sentences and all the previous generated target sentences are passed as context $C^{(i)}=\{x^{(1)}, ..., x^{(K)}, y^{(1)}, ..., y^{(i-1)}\}$. 

As mentioned, there are many architectural approaches to leveraging context (see \S\ref{related-work} for a more complete review), and the methods that we present in this paper are compatible with most architectures because they do not specify how the model $q_\theta$ uses the context. In experiments, we focus mostly on the simpler approach of concatenating the context to the current sentences \cite{tiedemann-scherrer-2017-neural}. 
Recent work by \citet{lopes-etal-2020-document} has shown that, given enough data (either through pre-training or larger contextual datasets), this simple approach tends to be competitive with or even outperform its more complex counterparts 

\section{Measuring Context Usage}\label{sec:xmi}

\subsection{Conditional Cross-Mutual Information}

While context-aware models \emph{allow} use of context, they \emph{do not ensure} contextual information is actually used: models could just be relying on the current source sentence and/or previously generated target words from the same sentence when generating the output.

Contrastive evaluation, where models are assessed based on the ability to distinguish correct translations from contrastive ones, is a common way to assess the ability of context-aware models to capture specific discourse phenomena that require inter-sentential context, such as anaphora resolution \cite{muller-etal-2018-large} and lexical cohesion \cite{bawden-etal-2018-evaluating}.
However, these methods only provide an indirect measure of context usage with respect to a limited number of phenomena and can fail to capture other, unknown ways in which the model might be using context. \citet{kim-etal-2019-document} showed that most improvements to translation quality are due to non-interpretable usages of context, such as the introduction of noise that acts as a regularizer to the encoder/decoder. This problem is further exacerbated by the fact that there is no clear definition of what entails ``context usage''.

In a different context, \citet{bugliarello-etal-2020-easier} introduced \textit{cross-mutual information} (XMI), to measure the ``difficulty'' of translating between different language pairs in sentence-level neural machine translation. Given a language model $q_{LM}$ for a target sentence $Y$ and a translation model $q_{MT}$ for translating from $X$ to $Y$, XMI is defined as:
$$
\text{XMI}(X\rightarrow Y) = \text{H}_{q_{LM}}(Y) - \text{H}_{q_{MT}}(Y|X), 
$$
where $\text{H}_{q_{LM}}$ denotes the cross-entropy of the target sentence $Y$ under the language model $q_{LM}$ and $\text{H}_{q_{MT}}$ the conditional cross-entropy of $Y$ given $X$ under the translation model $q_{MT}$. This allows us to measure how much information the source sentence gives us about the target sentence (an analogue of mutual information for cross-entropy). In the case where $q_{LM}$ and $q_{MT}$ perfectly model the underlying probabilities we would have $\text{XMI}(X\rightarrow Y) = \text{MI}(X, Y)$, the true mutual information.


Taking inspiration from the above, we propose \textbf{Conditional Cross-Mutual Information} (\textsc{CXMI}), a new measure of the influence of context on a model's predictions. This is done by considering an additional variable for the context $C$ and measuring how much information the context $C$ provides about the target $Y$ given the source $X$. This can then be formulated as 
\begin{align*}
\text{\textsc{CXMI}}(C\rightarrow Y|X) &=   \\
& \hspace{-2em} \text{H}_{q_{MT_A}}(Y|X) - \text{H}_{q_{MT_C}}(Y|X,C)
\end{align*}
where $\text{H}_{q_{MT_A}}$ is the entropy of a  \textit{context-agnostic} machine translation model,  and $\text{H}_{q_{MT_C}}$ refers to a  \textit{context-aware} machine translation model. This quantity can be estimated (see Appendix \ref{estimating-cxmi} for a more formal derivation) over an held-out test set with $N$ sentence pairs and the respective context as:
\begin{align*}
\text{\textsc{CXMI}}(C\rightarrow Y|X) & \approx \\
& \hspace{-2.2em} -\frac{1}{N} \sum_{i=1}^N \log \frac{q_{MT_A}(y^{(i)}|x^{(i)})}{q_{MT_C}(y^{(i)}|x^{(i)},C^{(i)})}
\end{align*}
While $q_{MT_A}$ and $q_{MT_C}$ can, in theory, be any models, we are interested in removing any confounding factors other than the context that might lead to instability in the estimates of the distributions. For example, if $q_{MT_A}$ and $q_{MT_C}$ use completely different models, it would not be clear if the difference in the probability estimates is due to the introduction of context or due to other extraneous factors such as differences in architectures, training regimens, or random seeds. To address this we consider a single model, $q_{MT}$, that is able to translate with and without context (more on how this achieved in \S\ref{section:xmi-experiments}). We can then set the context-agnostic model and the contextual model to be the same model $q_{MT_A}=q_{MT_C}=q_{MT}$. This way we attribute the information gain to the introduction of context. Throughout the rest of this work, when we reference ``context usage'' we will precisely mean this information gain (or loss).

\subsection{Experiments}
\label{section:xmi-experiments}

\paragraph{Data}
We experiment with a document-level translation task by training models on the IWSLT2017 \cite{cettolo-2012} dataset for language pairs $\text{EN}\rightarrow \text{DE}$ and $\text{EN}\rightarrow \text{FR}$ (with approximately 200K sentences for both pairs). We use the test sets 2011-2014 as validation sets and the 2015 as test sets.  To address the concerns pointed out by \citet{lopes-etal-2020-document} that gains in performance are due to the use of small training corpora and weak baselines, we use Paracrawl \cite{espla-etal-2019-paracrawl} and perform some data cleaning based on language identification tools, creating a pretraining dataset of around 82M and 104M sentence pairs for $\text{EN}\rightarrow \text{DE}$ and $\text{EN}\rightarrow \text{FR}$ respectively.

All data is encoded/vectorized with byte-pair encoding \cite{sennrich-etal-2016-neural} using the \textit{SentencePiece} framework \cite{kudo-richardson-2018-sentencepiece}. For the non-pretrained case, we use 20K vocabulary size shared across source/target, while for the pretrained case we use a 32K vocabulary size.

Besides translation quality, we also evaluate our models on two contrastive datasets for different discourse phenomena to better assess the ability of our models to capture context (more on this in \S\ref{sec:source-dropout-results}):
\begin{itemize}
    \item For the $\text{EN}\rightarrow\text{DE}$ language pair, we evaluate on the \textit{ContraPro} dataset \cite{muller-etal-2018-large}, targeting anaphoric pronoun resolution. Source-side sentences contain the English anaphoric pronoun \textit{it} while target-side sentences contain the corresponding German translations \textit{er}, \textit{sie} or \textit{es}. Contrastive erroneous translations are automatically created by replacing the correct pronoun with one of the other two. The test set contains 4,000 examples for each target pronoun type and context is needed to correctly disambiguate. Context includes the four previous sentences
    \item For the $\text{EN}\rightarrow\text{FR}$ language pair, we evaluate on the dataset by \citet{bawden-etal-2018-evaluating} targeting anaphoric pronoun resolution and lexical cohesion. It contains 200 manually curated examples for each phenomenon. Anaphora examples include singular and plural personal and possessive pronouns that require context to be correctly inferred and the dataset is balanced such that a model that does not use context can only achieve 50\% accuracy. Context includes the previous sentence
\end{itemize}

\paragraph{Models and Optimization}

For all our experiments, we consider an encoder-decoder Transformer architecture \cite{46201}. In particular, we train the \textit{transformer small} (hidden size of 512, feedforward size of 1024, 6 layers, 8 attention heads). For the pretrained setup, we also pre-train a \textit{transformer large} architecture (hidden size of 1024, feedforward size of 4096, 6 layers, 16 attention heads) and subsequently fine-tune on the IWSL2017 datasets.

As in \citet{46201}, we train using the Adam optimizer with $\beta_1=0.9$ and $\beta_2=0.98$ and use an inverse square root learning rate scheduler, with an initial value of $10^{-4}$ and $5\times 10^{-4}$ for pretrained and non-pretrained cases respectively, and with a linear warm-up in the first $4000$ steps. We train the models with early stopping on the validation perplexity. 

We train all our models on top of the \textit{Fairseq} framework \cite{ott-etal-2019-fairseq}. 

\paragraph{What Context Matters?\\}
\label{sec:importance-context}

\begin{figure*}[h]
\centering
\begin{tikzpicture}
\begin{axis}[
	xlabel=Context Size,
	x label style={at={(axis description cs:0.5,-0.1)},anchor=north},
    y label style={at={(axis description cs:-0.11,.5)},anchor=south},
	ylabel=\textsc{CXMI},
	xtick={0,1,2,3,4},
	ymin=-1e-2, ymax=2.5e-2,
	width=8cm,height=6cm,
    legend pos=south west
]
\addplot[color=red,mark=square*] coordinates {
    (0, 0)
	(1, 0.0078)
	(2, 0.0115)
	(3, 0.0076)
	(4, -0.0037)
};

\addplot[color=blue,mark=*] coordinates {
    (0, 0)
	(1, 0.0117)
	(2, 0.0144)
	(3, 0.0150)
	(4, 0.0151)
};
\legend{Source, Target}
\end{axis}
\end{tikzpicture}
\begin{tikzpicture}
\begin{axis}[
	xlabel=Context Size,
	x label style={at={(axis description cs:0.5,-0.1)},anchor=north},
    y label style={at={(axis description cs:-0.11,.5)},anchor=south},
	xtick={0,1,2,3,4},
	ymin=-1e-2, ymax=2.5e-2,
	width=8cm,height=6cm,
    legend pos=south west
]
\addplot[color=red,mark=square*] coordinates {
    (0, 0)
	(1, 0.0016)
	(2, -0.0003)
	(3, -0.0013)
	(4, -0.0049)
};
\addplot[color=blue,mark=*] coordinates {
    (0, 0)
	(1, 0.0212)
	(2, 0.0214)
	(3, 0.0219)
	(4, 0.0197)
};
\legend{Source, Target}
\end{axis}
\end{tikzpicture}
\caption{$\text{\textsc{CXMI}}$ values for the $\text{EN}\rightarrow \text{DE}$ as a function of source and target context sizes for non-pretrained (left) and pretrained (right) models.}
\label{fig:xmi-context-size-general}
\end{figure*}
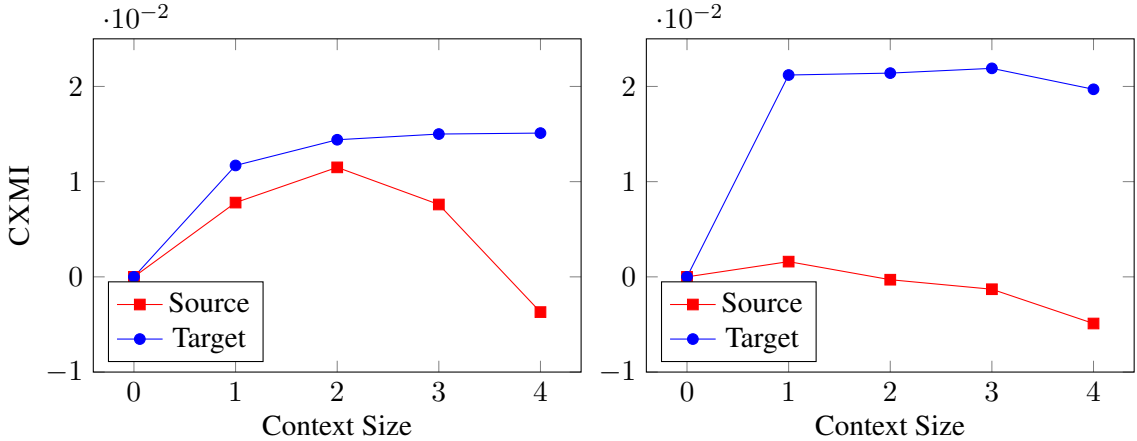

To assess the relative importance of different context sizes on both the source and target side, we start by considering two models, one for the source-side context and one for the target-side context, that receive context of size $k$, $C^{(i)}=\{x^{(i-k)},\ldots,x^{(i-1)}\}$ or $C^{(i)}=\{y^{(i-k)},\ldots,y^{(i-1)}\}$. During training, $k$ is selected randomly to be in $\{1,\ldots,4\}$ for every example. This way the model is trained to translate the same source without and with different context sizes and is thus able to translate based on any context size in that interval. 

Figure \ref{fig:xmi-context-size-general} shows the \textsc{CXMI} values computed over the test set as a function of the context size for both the source-side and target-side contextual models for both the non-pretrained and pretrained regimens for the $\text{EN}\rightarrow \text{DE}$ language pair. Results for the $\text{EN}\rightarrow \text{FR}$ language pair are similar and can be found in Appendix \ref{appendix:cxmi-enfr}.

For the non-pretrained case, for both the source and target context, the biggest jump in context usage is when we increase the context size from 0 to 1. After that, increasing the context size leads to diminishing increases in context usage and even \textit{reduced} context usage for the source-side context.
Interestingly, when the model is stronger, such as in the pretrained case, we can see that it can leverage target-side context even better than the non-pretrained case, with a similar trend of diminishing increases in context usage for both regimes. However, this is not the case for the source-side context, and it seems that the pretrained model is barely able to use the contextual information on this side.

Overall, for this regime, we can conclude that having a context size of one or two previous sentences on both sides is beneficial to the model, and that target-side context is \textit{slightly} more used than source-side context.  This appears to corroborate the findings of \citet{bawden-etal-2018-evaluating} that target-side context is more effective than the source context.

\pagebreak

\paragraph{Does \textsc{CXMI} Really Measure Context Usage?}

\begin{table}
\centering
\begin{tabular}{ cccc }
 \toprule
   & \multicolumn{3}{c}{$r_{pb}$} \\
   \cmidrule(lr){2-4}
  Context Size & (1) & (2) & (3) \\
 \midrule
  1 & \textbf{0.365} & \textbf{0.315} & \textbf{0.206}\\ 
  2 & \textbf{0.366} & - & -\\ 
  3 & \textbf{0.367} & - & -\\
  4 & \textbf{0.366} & - & -\\
 \hline
\end{tabular}
\caption{Point-Biserial correlation coefficients on the contrastive datasets with pretrained models for different context sizes. Measured on \textit{ContraPro} (1) and  \citet{bawden-etal-2018-evaluating}, both for pronoun resolution (2) and lexical cohesion (3). Bold values mean the correlation is statistically significant with $p < 0.01$.}
\label{cxmi-correlations}
\end{table}

To assert that \textsc{CXMI} correlates with interpretable measures of context usage, we perform a correlation analysis with the performance in the contrastive datasets mentioned. In these datasets, usage of context is evident where the model picks the right answer when it is passed the context and is not able to do so when no context is given. Thus Table \ref{cxmi-correlations} shows the point-biserial correlation coefficient%
\footnote{The Point-Biserial correlation coefficient is a special case of the \textit{Pearson} correlation coefficient when one of the random variables is dichotomous.} between the \textit{per-sample} \textsc{CXMI} and binary random variable and a binary variable that takes the value 1 if the contextual model picks the correct translation and the non-contextual model picks the incorrect one, for different context sizes on the pretrained model. We can see that there is a statistically significant correlation between both values, which strengthens the notion that \textsc{CXMI} captures previous measures of context usage to some extent.

\section{Increasing Context Usage}\label{sec:source_dropout}

\subsection{Context-aware Word Dropout}

Motivated by the above results demonstrating the limited context usage of models trained using the standard MLE training paradigm, particularly with respect to more distant context, we now ask the question: ``Is it possible to modify the training methodology to increase context usage by the model?'' As an answer, we extend a popular regularization technique used in sentence-level machine translation, word dropout \cite{sennrich-etal-2016-edinburgh}, to the context-aware setting. The idea behind \underline{co}ntext-aware \underline{word} (\textsc{CoWord}) dropout is to model the translation probability between $x^{(i)}$ and $y^{(i)}$ as
$$
p_\theta(y^{(i)} | x^{(i)}) = \prod_{t=1}^T p_\theta(y^{(i)}_t | \tilde x^{(i)}, y^{(i)}_{<t}, C^{(i)}),\\
$$
where $\tilde x^{(i)}$ is a perturbed version of the current source sentence generated by randomly dropping tokens and replacing them with a mask token given a dropout probability $p$:
\begin{align*}
    r^{(i)}_t &\sim \text{Bernoulli}(p) \\
    \tilde x^{(i)}_t &= 
    \begin{cases}
        \langle\textsc{mask}\rangle & \text{if  } r^{(i)}_t = 1 \\ 
        x^{(i)}_t & \text{otherwise.}
    \end{cases}   
\end{align*}

In the case where no context is passed $C^{(i)} = \varnothing$, \textsc{CoWord} dropout reduces to word dropout.
The intuition behind such a perturbation is that, by dropping information from the current source and \emph{not} the context, we increase the relative reliability of context $C^{(i)}$, therefore providing the inductive bias that context is important for the translation. We will see in \S\ref{sec:experiments} that this inductive bias is beneficial and that \textsc{CoWord} dropout not only improves performance but also increases context usage. 

\subsection{Experiments}
\label{sec:experiments}
\paragraph{Setup} As in \S\ref{section:xmi-experiments}, we consider \textit{transformer} models trained on the IWSLT2017 for both $\text{EN}\rightarrow \text{DE}$ and $\text{EN}\rightarrow \text{FR}$, both from scratch and pretrained using the procedure previously described. In particular, due to findings in the previous section, we consider models with either only \textit{target-side} context or both \textit{source-side} and \textit{target-side} context.

\paragraph{Context Usage}

\label{sec:source-dropout-results}

To assess if our proposed regularization technique, \textsc{CoWord} dropout, increases context usage by models, we train a model using the same \textit{dynamic} context size setting used in \S\ref{sec:importance-context}. 

\begin{figure}
\begin{tikzpicture}
\begin{axis}[
	xlabel=Context Size,
	x label style={at={(axis description cs:0.5,-0.1)},anchor=north},
    y label style={at={(axis description cs:-0.11,.5)},anchor=south},
	ylabel=\textsc{CXMI},
	xtick={0,1,2,3,4},
	width=8cm,height=6cm,
    legend pos=south east
]
\addplot[color=blue!40, mark=x] coordinates {
    (0, 0)
	(1, 0.0117)
	(2, 0.0144)
	(3, 0.0150)
	(4, 0.0151)
};

\addplot[color=blue!60,mark=triangle*] coordinates {
    (0, 0)
	(1, 0.0160)
	(2, 0.0183)
	(3, 0.0191)
	(4, 0.0194)
};
\addplot[color=blue!80,mark=diamond*] coordinates {
    (0, 0)
	(1, 0.0186)
	(2, 0.0199)
	(3, 0.0204)
	(4, 0.0204)
};
\addplot[color=blue,mark=*] coordinates {
    (0, 0)
	(1, 0.0210)
	(2, 0.0259)
	(3, 0.0271)
	(4, 0.0265)
};
\legend{$p=0.0$, $p=0.1$, $p=0.2$, $p=0.3$}
\end{axis}
\end{tikzpicture}
\caption{$\text{\textsc{CXMI}}$ values as a function target context size for different values of \textsc{CoWord} dropout}
\label{fig:xmi-context-size-source-dropout}
\end{figure}
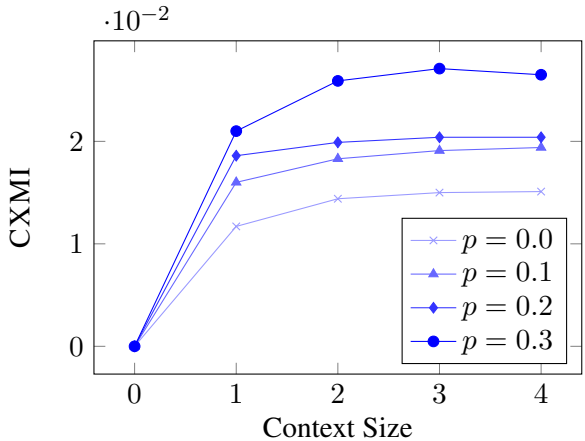

Figure \ref{fig:xmi-context-size-source-dropout} plots the \textsc{CXMI} values on the test set as a function of the \textit{target} context size as we increase the dropout value $p$. We see that increasing this value consistently increases context usage according to \textsc{CXMI} across different context sizes. Note that, at test time, \textsc{CoWord} dropout is disabled, which means that it provides inductive bias only during training and models \emph{learn} to use more context by themselves.

\setulcolor{red} 
\begin{table*}[ht]
\scalebox{0.85}{
\hspace{1em}
\begin{tabular}{ p{0.21\textwidth}|p{0.21\textwidth}|p{0.2\textwidth}|p{0.19\textwidth} |m{0.10\textwidth}}
 \toprule
  \textbf{Source Context} & \textbf{Source} & \textbf{Target Context} & \textbf{Target} & $\Delta \text{\textsc{CXMI}}$  \\
  \midrule
    More people watched games because it was faster. & It was more entertaining & \hlc[blue!30]{Mehr Menschen sahen zu, die Spiele wurden schneller} & \dunderline{2pt}{\hlc[green!50]{und}}\hlc[green!50]{ unterhaltsamer.} & 0.53 \\
  \hline
  The ball comes off track. & You don't know where it's going to land & \hlc[blue!30]{Der Ball} ist außer Kontrolle & Sie wissen nicht, wo \dunderline{2pt}{\hlc[green!50]{er}} landet. & 0.33 \\
   \hline
  I really think that this lie that we've been sold about disability is the greatest injustice & It makes life hard for us & Meiner Meinung nach ist diese \hlc[blue!30]{Luge} über Behinderung eine schreiende Ungerechtigkeit & \dunderline{2pt}{\hlc[green!50]{Sie}} macht uns das Leben schwer. & 0.25\\
 \bottomrule
\end{tabular}
}
\caption{Examples where models with \textsc{CoWord} dropout use the target context more than models trained without it. Word highlighted blue in the context are used to disambiguate translations while highlighted green in the target use context according to native speakers. Words underlined in the target are the ones with the highest \textit{per-word} CXMI i.e. the ones that use the most context according to the model}
\label{table:cxmi-examples}
\end{table*}

Table \ref{table:cxmi-examples} illustrates some examples where the \textsc{CoWord} dropout increased the \textit{per-sample} \textsc{CXMI} significantly. While the model only has access to \textit{target} context, we present the source context for clarity. In the first example, while the source is a complete sentence, the target is only a fragment of one so the context helps complete it. In the other two examples shown, we can see that context helps disambiguate the gender of the German translation of the English pronoun \textit{it}. Interestingly, the words that use context the most according to CXMI match very closely to the ones that native speakers annotated. 

\paragraph{Translation Quality}

\begin{table*}[h!]
\centering
\scalebox{0.92}{
\begin{tabular}{ cccccccccc } 
\toprule
& & \multicolumn{4}{c}{$\text{EN}\rightarrow \text{DE}$} & \multicolumn{4}{c}{$\text{EN}\rightarrow \text{FR}$} \\
\cmidrule(lr){3-6} \cmidrule(lr){7-10}
& & \multicolumn{2}{c}{} & \multicolumn{2}{c}{w/ pretraining} & \multicolumn{2}{c}{} & \multicolumn{2}{c}{w/ pretraining}\\
\cmidrule(lr){3-4} \cmidrule(lr){5-6} \cmidrule(lr){7-8} \cmidrule(lr){9-10}
& $p$ & BLEU & COMET & BLEU & COMET & BLEU & COMET & BLEU & COMET\\
\midrule
\multirow{3}{*}{baseline} &0.0 &26.36 &0.083 &35.10 &0.521 &37.62 &0.450 &42.98 &0.679 \\
&0.1 &\textbf{27.26} &0.159 &\textbf{35.15} &\textbf{0.525} &38.16 &0.472 &\textbf{43.28} &\textbf{0.679} \\
&0.2 &26.97 &\textbf{0.163} &35.13 &0.524 &\textbf{38.34} &\textbf{0.474} &42.99 &0.678 \\
\midrule
\multirow{3}{*}{1-to-2} &0.0 &26.60 &0.087 &\textbf{35.22} &\textbf{0.528} &37.59 &0.450 &42.89 &0.672 \\
&0.1 &\textbf{27.36} &0.174 &34.92 &0.527 &38.25 &0.472 &42.88 &0.677 \\
&0.2 &27.33 &\textbf{0.193} &34.75 &0.524 &\textbf{38.27} &\textbf{0.485} &\textbf{42.90} &\textbf{0.678} \\
\midrule
\multirow{3}{*}{2-to-2} &0.0 &26.85 &0.090 &34.47 &0.471 &37.54 &0.453 &\textbf{42.97} &0.674 \\
&0.1 &\textbf{27.72} &0.169 &34.51 &0.522 &\textbf{38.30} &0.467 &42.95 &\textbf{0.676} \\
&0.2 &27.21 &\textbf{0.177} &\textbf{34.65} &\textbf{0.525} &38.15 &\textbf{0.468} &42.88 &0.675 \\
\bottomrule
\end{tabular}
}
\caption{\label{table:source-dropout-iwslt}Results on IWSLT2017 with different probabilities for \textsc{CoWord} dropout. Averaged across three runs for each method.}
\end{table*}

\begin{table}[h!]
\centering
\resizebox{\linewidth}{!}{
\begin{tabular}{ cccccc } 
\toprule
& & \multicolumn{2}{c}{$\text{EN}\rightarrow \text{DE}$} & \multicolumn{2}{c}{$\text{EN}\rightarrow \text{FR}$} \\
\cmidrule(lr){3-4} \cmidrule(lr){5-6} 
& $p$ & BLEU & COMET & BLEU & COMET \\
\midrule
\multirow{3}{*}{baseline} &0.0 &26.36 &0.083 &37.62 &0.450 \\
&0.1 &\textbf{27.26} &0.159 &38.16 &0.472 \\
&0.2 &26.97 &\textbf{0.163} &\textbf{38.34} &\textbf{0.474} \\
\midrule
\multirow{3}{*}{multi} &0.0 &26.64 &0.104 &37.85 &0.466 \\
&0.1 &\textbf{27.45} &0.190 &37.98 &0.460 \\
&0.2 &27.31 &\textbf{0.190} &\textbf{38.30} &\textbf{0.484} \\
\bottomrule
\end{tabular}
}
\caption{\label{table:source-dropout-multi-iwslt}Results on IWSLT2017 for a multi-encoder 1-to-2 model with different probabilities for \textsc{CoWord} dropout. Averaged across three runs for each method.}
\vspace{-4mm}
\end{table}

To evaluate if the increased usage of context correlates with better machine translation quality, based on the previous experiments on context usage and values for \textsc{CoWord} dropout, we consider three models trained with \textit{fixed-size} context:
\begin{itemize}
    \item A \textbf{baseline} that has no context, reducing to sentence-level model ie: \textit{i.e.}, $C^{(i)} = \varnothing$;
    \item a \textbf{one-to-two} model having as context the previous target sentence, \textit{i.e.},  $C^{(i)}=\{y^{(i-1)}\}$;
    \item a \textbf{two-to-two} model having as context the previous source sentence and the previous target sentence, \textit{i.e.}, $C^{(i)}=\{x^{(i-1)}, y^{(i-1)}\}$.
\end{itemize}

In addition, to explore the benefits of  \textsc{CoWord} dropout in other architectures, we also train a \textbf{one-to-two} \textit{multi-encoder}  \cite{jean2017does} \textit{transformer small} model (more details in Appendix \S\ref{appendix:multiencoder}). For all models with target context, when decoding, we use the previous decoded sentences as target context. 

Table \ref{table:source-dropout-iwslt} shows the performance across three different seeds of the baseline and contextual models for both the non-pretrained and pretrained setting, with increasing values of \textsc{CoWord} dropout $p$. We also run the baseline with \textsc{CoWord} dropout (which, as said previously, reduces to word dropout) to ensure that improvements were not only due to regularization effects on the \textit{current} source/target. We report the standard BLEU score \cite{papineni2002bleu} calculated using sacreBLEU \cite{post-2018-call} and COMET, a more accurate evaluation method using multilingual embeddings \cite{rei-etal-2020-comet}. 

For the non-pretrained case, we can see that a \textsc{CoWord} dropout value $p > 0$ consistently improves the performance of the contextual models when compared to models running with $p=0$ and with the sentence-level baseline with the same values for word dropout. For the pretrained case, the improvements are not as noticeable, although models trained with \textsc{CoWord} dropout still always outperform models trained without it. This is perhaps a reflection of the general trend that better models are harder to improve.

\pagebreak

Table \ref{table:source-dropout-multi-iwslt} shows that \textsc{CoWord} dropout is also helpful for the \textit{multi-encoder} model, with \textsc{CoWord} dropout helping significantly. This shows that this method could be helpful for context-aware architectures other than concatenation-based.

\begin{table*}[hbt!]
\centering
\begin{tabular}{ ccccccccc } 
\toprule
& & \multicolumn{2}{c}{$\text{EN}\rightarrow \text{DE}$} & \multicolumn{4}{c}{$\text{EN}\rightarrow \text{FR}$} \\
\cmidrule(lr){3-4} \cmidrule(lr){5-8}
& & & w/ pretraining & \multicolumn{2}{c}{} & \multicolumn{2}{c}{w/ pretraining}\\
\cmidrule(lr){3-3} \cmidrule(lr){4-4} \cmidrule(lr){5-6} \cmidrule(lr){7-8}
& $p$ & Pronouns & Pronouns & Pronouns & Cohesion & Pronouns & Cohesion \\
\midrule
baseline &0.0 &42.96 &46.57 &50.00 &50.00 &50.00 &50.00 \\
\midrule
\multirow{3}{*}{1-to-2} &0.0 &57.36 &76.79 &68.16 &49.99 &\textbf{86.83} &\textbf{56.83} \\
&0.1 &58.70 &76.28 &72.33 &51.49 &86.49 &56.66 \\
&0.2 &\textbf{60.72} &\textbf{77.52} &\textbf{72.99} &\textbf{52.16} &85.66 &56.49 \\
\midrule
\multirow{3}{*}{2-to-2} &0.0 &61.06 &80.33 &72.16 &50.99 &85.66 &64.33 \\
&0.1 &\textbf{66.00} &\textbf{80.35} &\textbf{73.99} &\textbf{52.49} &87.16 &\textbf{65.99} \\
&0.2 &65.47 &79.97 &\textbf{73.99} &\textbf{52.49} &\textbf{88.49} &63.99 \\
\bottomrule
\end{tabular}
\caption{\label{table:contrastive-datasets} Results on anaphoric pronoun resolution and lexical cohesion contrastive datasets with different probabilities for \textsc{CoWord} dropout. Averaged across three runs for each method.}%
\end{table*}

\begin{table}[hbt!]
\centering
\resizebox{\linewidth}{!}{
\begin{tabular}{ ccccc } 
\toprule
& & $\text{EN}\rightarrow \text{DE}$ & \multicolumn{2}{c}{$\text{EN}\rightarrow \text{FR}$} \\
\cmidrule(lr){3-3} \cmidrule(lr){4-5} 
& $p$ & Pronouns & Pronouns & Cohesion  \\
\midrule
baseline &0.0 &42.96 &50.00 &50.00 \\
\midrule
\multirow{3}{*}{multi} &0.0 &42.85 &49.74 &49.99 \\
&0.1 &47.29 &51.74 &50.24 \\
&0.2 &\textbf{47.57} &\textbf{52.50} &\textbf{50.99} \\
\bottomrule
\end{tabular}
}
\caption{\label{table:contrastive-datasets-multi} Results on anaphoric pronoun resolution and lexical cohesion contrastive datasets for the multi-encoder 1-to-2 model with different probabilities for \textsc{CoWord} dropout. Averaged across three runs for each method.}
\end{table}

\paragraph{Discourse Phenomena}

While automatic metrics such as BLEU and COMET allow us to measure translation quality, they mostly target \textit{sentence-level} quality and do not specifically focus on phenomena that require context-awareness. Contrastive datasets, as described in \S\ref{section:xmi-experiments}, allow us to measure the performance of context-aware models in specific discourse phenomena by comparing the probability of \textit{correct} translation against the \textit{contrastive} translations. Models that capture the targeted discourse phenomena well will consistently rank the correct translation higher than the contrastive ones. While there is a disconnect between the translation (done via decoding) and contrastive evaluation, it is currently the best way to measure a model's performance on context-aware discourse phenomena.

\pagebreak

Table \ref{table:contrastive-datasets} shows the average performance over the contrastive datasets of the baseline and contextual models for both the (non-)pretrained settings, with increasing values of \textsc{CoWord} dropout $p$. We can see that in general, increasing \textsc{CoWord} dropout leads to improved performance, particularly for the non-pretrained case. This gain is particularly clear for pronoun resolution and the $\text{EN}\rightarrow \text{DE}$ language pair. We hypothesise that this is due to the small size of the contrastive sets for the $\text{EN}\rightarrow \text{FR}$ language pair, which leads to high variance.

Table \ref{table:contrastive-datasets-multi} similarly shows that \textsc{CoWord} dropout improves the performance of the multi-encoder model across all phenomena, which again shows that our proposed regularization method has benefits for multiple architectures for context-aware machine translation. Curiously, when these models are trained without \textsc{CoWord} dropout, they achieve performance similar to the sentence-level baseline, while when dropout is applied, they are able to effectively start using context.

\pagebreak

\section{Related Work} \label{related-work}

\paragraph{Context-aware Machine Translation}
There have been many works in the literature that try to incorporate context into NMT systems. \citet{tiedemann-scherrer-2017-neural} first proposed the simple approach of concatenating the previous sentences in both the source and target side to the input to the system; \citet{jean2017does}, \citet{bawden-etal-2018-evaluating}, and \citet{zhang-etal-2018-improving} used an additional context-specific encoder to extract contextual features from the previous sentences; \citet{maruf-haffari-2018-document} and \citet{Tu2018LearningTR} used cache-based memories to encode context; \citet{wang-etal-2017-exploiting-cross} used a hierarchical RNN to encode the global context from all previous sentences; \citet{miculicich-etal-2018-document} and \citet{maruf-etal-2019-selective} used hierarchical attention networks to encode context; \citet{chen-etal-2020-modeling-discourse} added document-level discourse structure information to the input;  \citet{sun2020document} trained a simple concatenation-based model with \textit{varying} context size during training to have a model that is able to translate with any context size, similar to what is done in this work. Similarly to what we do with \textsc{CoWord} dropout, \citet{jean-cho-2019} attempted to maximise sensitivity to context by introducing a margin-based regularization term to explicitly encourage context usage.

For a more detailed overview, \citet{Maruf2019ASO} extensively describe the different approaches and how they leverage context. While these models lead to improvements with small training sets, \citet{lopes-etal-2020-document} showed that the improvements are negligible when compared with the concatenation baseline when using larger datasets.
However, importantly, both our metric \textsc{CXMI} for measuring context usage and the proposed regularization method of \textsc{CoWord} dropout, can theoretically be applied to any of the above-mentioned methods. 

\paragraph{Evaluation}
In terms of evaluation, most previous work focuses on targeting a system's performance on contrastive datasets for specific inter-sentential discourse phenomena. \citet{muller-etal-2018-large} built a large-scale dataset for anaphoric pronoun resolution, \citet{bawden-etal-2018-evaluating} manually created a dataset for both pronoun resolution and lexical choice and \citet{voita-etal-2019-good} created a dataset that targets deixis, ellipsis and lexical cohesion. \citet{stojanovski-etal-2020-contracat} showed through adversarial attacks that models that do well on other contrastive datasets rely on surface heuristics and create a contrastive dataset to address this.
In contrast, our \textsc{CXMI} metric is phenomenon-agnostic and can be measured with respect to \emph{all} phenomena that require context in translation.

\paragraph{Information-Theoretic Analysis}
\citet{bugliarello-etal-2020-easier} first proposed cross-mutual information (XMI) in the context of measuring the difficulty of translating between languages. Our work differs in that we propose a \textit{conditional} version of XMI, where $S$ is always observed, and we use it to assess the information gain of context rather than the difficulty of translating different languages.

\section{Implications and Future Work}

We introduce a new, architecture-agnostic, metric to measure how context-aware machine translation models are using context and propose a simple regularization technique to increase context usage by these models. Our results are theoretically applicable to almost all recently proposed context-aware models and future work should go about measuring exactly how much these models leverage context and if \textsc{CoWord} dropout also improves context usage and performance in these.

We also hope this work motivates exploring (C)XMI for other uses cases where measuring the relevance/usage of inputs to a particular model other than context-aware machine translation. It could, for example, be used in conditional language modelling to analyse how the inputs we are conditioning on are being used by the model.

\section{Acknowledgements}

We would like to thank all the members of DeepSPIN, NeuLab, and Unbabel who provided feedback on earlier versions of this work. This work was 
supported by the European Research Council (ERC StG DeepSPIN 758969),
by the P2020 programs MAIA and Unbabel4EU (LISBOA-01-0247-FEDER-045909 and LISBOA-01-0247-FEDER-042671), and by the Funda\c{c}\~ao para a Ci\^encia e Tecnologia through contracts SFRH/BD/150706/2020 and UIDB/50008/2020.


\bibliographystyle{acl_natbib}
\bibliography{anthology,acl2021}

\begin{thebibliography}{33}
\expandafter\ifx\csname natexlab\endcsname\relax\def\natexlab#1{#1}\fi

\bibitem[{Bawden et~al.(2018)Bawden, Sennrich, Birch, and
  Haddow}]{bawden-etal-2018-evaluating}
Rachel Bawden, Rico Sennrich, Alexandra Birch, and Barry Haddow. 2018.
\newblock \href {https://doi.org/10.18653/v1/N18-1118} {Evaluating discourse
  phenomena in neural machine translation}.
\newblock In \emph{Proceedings of the 2018 Conference of the North {A}merican
  Chapter of the Association for Computational Linguistics: Human Language
  Technologies, Volume 1 (Long Papers)}, pages 1304--1313, New Orleans,
  Louisiana. Association for Computational Linguistics.

\bibitem[{Bugliarello et~al.(2020)Bugliarello, Mielke, Anastasopoulos,
  Cotterell, and Okazaki}]{bugliarello-etal-2020-easier}
Emanuele Bugliarello, Sabrina~J. Mielke, Antonios Anastasopoulos, Ryan
  Cotterell, and Naoaki Okazaki. 2020.
\newblock \href {https://doi.org/10.18653/v1/2020.acl-main.149} {It{'}s easier
  to translate out of {E}nglish than into it: {M}easuring neural translation
  difficulty by cross-mutual information}.
\newblock In \emph{Proceedings of the 58th Annual Meeting of the Association
  for Computational Linguistics}, pages 1640--1649, Online. Association for
  Computational Linguistics.

\bibitem[{Cettolo et~al.(2012)Cettolo, Girardi, and Federico}]{cettolo-2012}
Mauro Cettolo, Christian Girardi, and Marcello Federico. 2012.
\newblock \href {https://www.aclweb.org/anthology/2012.eamt-1.60} {{WIT}3: Web
  inventory of transcribed and translated talks}.
\newblock In \emph{Proceedings of the 16th Annual conference of the European
  Association for Machine Translation}, pages 261--268, Trento, Italy. European
  Association for Machine Translation.

\bibitem[{Chen et~al.(2020)Chen, Li, Zhang, Zhou, Cui, Wang, and
  Su}]{chen-etal-2020-modeling-discourse}
Junxuan Chen, Xiang Li, Jiarui Zhang, Chulun Zhou, Jianwei Cui, Bin Wang, and
  Jinsong Su. 2020.
\newblock \href {https://doi.org/10.18653/v1/2020.autosimtrans-1.5} {Modeling
  discourse structure for document-level neural machine translation}.
\newblock In \emph{Proceedings of the First Workshop on Automatic Simultaneous
  Translation}, pages 30--36, Seattle, Washington. Association for
  Computational Linguistics.

\bibitem[{Espl{\`a} et~al.(2019)Espl{\`a}, Forcada, Ram{\'\i}rez-S{\'a}nchez,
  and Hoang}]{espla-etal-2019-paracrawl}
Miquel Espl{\`a}, Mikel Forcada, Gema Ram{\'\i}rez-S{\'a}nchez, and Hieu Hoang.
  2019.
\newblock \href {https://www.aclweb.org/anthology/W19-6721} {{P}ara{C}rawl:
  Web-scale parallel corpora for the languages of the {EU}}.
\newblock In \emph{Proceedings of Machine Translation Summit XVII Volume 2:
  Translator, Project and User Tracks}, pages 118--119, Dublin, Ireland.
  European Association for Machine Translation.

\bibitem[{Hassan et~al.(2018)Hassan, Aue, Chen, Chowdhary, Clark, Federmann,
  Huang, Junczys{-}Dowmunt, Lewis, Li, Liu, Liu, Luo, Menezes, Qin, Seide, Tan,
  Tian, Wu, Wu, Xia, Zhang, Zhang, and
  Zhou}]{DBLP:journals/corr/abs-1803-05567}
Hany Hassan, Anthony Aue, Chang Chen, Vishal Chowdhary, Jonathan Clark,
  Christian Federmann, Xuedong Huang, Marcin Junczys{-}Dowmunt, William Lewis,
  Mu~Li, Shujie Liu, Tie{-}Yan Liu, Renqian Luo, Arul Menezes, Tao Qin, Frank
  Seide, Xu~Tan, Fei Tian, Lijun Wu, Shuangzhi Wu, Yingce Xia, Dongdong Zhang,
  Zhirui Zhang, and Ming Zhou. 2018.
\newblock \href {http://arxiv.org/abs/1803.05567} {Achieving human parity on
  automatic chinese to english news translation}.
\newblock \emph{CoRR}, abs/1803.05567.

\bibitem[{Jean and Cho(2019)}]{jean-cho-2019}
S{\'{e}}bastien Jean and Kyunghyun Cho. 2019.
\newblock \href {http://arxiv.org/abs/1903.04715} {Context-aware learning for
  neural machine translation}.
\newblock \emph{CoRR}, abs/1903.04715.

\bibitem[{Jean et~al.(2017)Jean, Lauly, Firat, and Cho}]{jean2017does}
Sebastien Jean, Stanislas Lauly, Orhan Firat, and Kyunghyun Cho. 2017.
\newblock \href {http://arxiv.org/abs/1704.05135} {Does neural machine
  translation benefit from larger context?}

\bibitem[{Kim et~al.(2019)Kim, Tran, and Ney}]{kim-etal-2019-document}
Yunsu Kim, Duc~Thanh Tran, and Hermann Ney. 2019.
\newblock \href {https://doi.org/10.18653/v1/D19-6503} {When and why is
  document-level context useful in neural machine translation?}
\newblock In \emph{Proceedings of the Fourth Workshop on Discourse in Machine
  Translation (DiscoMT 2019)}, pages 24--34, Hong Kong, China. Association for
  Computational Linguistics.

\bibitem[{Kudo and Richardson(2018)}]{kudo-richardson-2018-sentencepiece}
Taku Kudo and John Richardson. 2018.
\newblock \href {https://doi.org/10.18653/v1/D18-2012} {{S}entence{P}iece: A
  simple and language independent subword tokenizer and detokenizer for neural
  text processing}.
\newblock In \emph{Proceedings of the 2018 Conference on Empirical Methods in
  Natural Language Processing: System Demonstrations}, pages 66--71, Brussels,
  Belgium. Association for Computational Linguistics.

\bibitem[{L{\"a}ubli et~al.(2018)L{\"a}ubli, Sennrich, and
  Volk}]{laubli-etal-2018-machine}
Samuel L{\"a}ubli, Rico Sennrich, and Martin Volk. 2018.
\newblock \href {https://doi.org/10.18653/v1/D18-1512} {Has machine translation
  achieved human parity? a case for document-level evaluation}.
\newblock In \emph{Proceedings of the 2018 Conference on Empirical Methods in
  Natural Language Processing}, pages 4791--4796, Brussels, Belgium.
  Association for Computational Linguistics.

\bibitem[{Lopes et~al.(2020)Lopes, Farajian, Bawden, Zhang, and
  Martins}]{lopes-etal-2020-document}
Ant{\'o}nio Lopes, M.~Amin Farajian, Rachel Bawden, Michael Zhang, and
  Andr{\'e} F.~T. Martins. 2020.
\newblock \href {https://www.aclweb.org/anthology/2020.eamt-1.24}
  {Document-level neural {MT}: A systematic comparison}.
\newblock In \emph{Proceedings of the 22nd Annual Conference of the European
  Association for Machine Translation}, pages 225--234, Lisboa, Portugal.
  European Association for Machine Translation.

\bibitem[{Maruf and Haffari(2018)}]{maruf-haffari-2018-document}
Sameen Maruf and Gholamreza Haffari. 2018.
\newblock \href {https://doi.org/10.18653/v1/P18-1118} {Document context neural
  machine translation with memory networks}.
\newblock In \emph{Proceedings of the 56th Annual Meeting of the Association
  for Computational Linguistics (Volume 1: Long Papers)}, pages 1275--1284,
  Melbourne, Australia. Association for Computational Linguistics.

\bibitem[{Maruf et~al.(2019{\natexlab{a}})Maruf, Martins, and
  Haffari}]{maruf-etal-2019-selective}
Sameen Maruf, Andr{\'e} F.~T. Martins, and Gholamreza Haffari.
  2019{\natexlab{a}}.
\newblock \href {https://doi.org/10.18653/v1/N19-1313} {Selective attention for
  context-aware neural machine translation}.
\newblock In \emph{Proceedings of the 2019 Conference of the North {A}merican
  Chapter of the Association for Computational Linguistics: Human Language
  Technologies, Volume 1 (Long and Short Papers)}, pages 3092--3102,
  Minneapolis, Minnesota. Association for Computational Linguistics.

\bibitem[{Maruf et~al.(2019{\natexlab{b}})Maruf, Saleh, and
  Haffari}]{Maruf2019ASO}
Sameen Maruf, Fahimeh Saleh, and Gholamreza Haffari. 2019{\natexlab{b}}.
\newblock A survey on document-level machine translation: Methods and
  evaluation.
\newblock \emph{ArXiv}, abs/1912.08494.

\bibitem[{Miculicich et~al.(2018)Miculicich, Ram, Pappas, and
  Henderson}]{miculicich-etal-2018-document}
Lesly Miculicich, Dhananjay Ram, Nikolaos Pappas, and James Henderson. 2018.
\newblock \href {https://doi.org/10.18653/v1/D18-1325} {Document-level neural
  machine translation with hierarchical attention networks}.
\newblock In \emph{Proceedings of the 2018 Conference on Empirical Methods in
  Natural Language Processing}, pages 2947--2954, Brussels, Belgium.
  Association for Computational Linguistics.

\bibitem[{M{\"u}ller et~al.(2018)M{\"u}ller, Rios, Voita, and
  Sennrich}]{muller-etal-2018-large}
Mathias M{\"u}ller, Annette Rios, Elena Voita, and Rico Sennrich. 2018.
\newblock \href {https://doi.org/10.18653/v1/W18-6307} {A large-scale test set
  for the evaluation of context-aware pronoun translation in neural machine
  translation}.
\newblock In \emph{Proceedings of the Third Conference on Machine Translation:
  Research Papers}, pages 61--72, Brussels, Belgium. Association for
  Computational Linguistics.

\bibitem[{Ott et~al.(2019)Ott, Edunov, Baevski, Fan, Gross, Ng, Grangier, and
  Auli}]{ott-etal-2019-fairseq}
Myle Ott, Sergey Edunov, Alexei Baevski, Angela Fan, Sam Gross, Nathan Ng,
  David Grangier, and Michael Auli. 2019.
\newblock \href {https://doi.org/10.18653/v1/N19-4009} {fairseq: A fast,
  extensible toolkit for sequence modeling}.
\newblock In \emph{Proceedings of the 2019 Conference of the North {A}merican
  Chapter of the Association for Computational Linguistics (Demonstrations)},
  pages 48--53, Minneapolis, Minnesota. Association for Computational
  Linguistics.

\bibitem[{Papineni et~al.(2002)Papineni, Roukos, Ward, and
  Zhu}]{papineni2002bleu}
Kishore Papineni, Salim Roukos, Todd Ward, and Wei-Jing Zhu. 2002.
\newblock Bleu: a method for automatic evaluation of machine translation.
\newblock In \emph{Proceedings of the 40th annual meeting of the Association
  for Computational Linguistics}, pages 311--318.

\bibitem[{Post(2018)}]{post-2018-call}
Matt Post. 2018.
\newblock \href {https://doi.org/10.18653/v1/W18-6319} {A call for clarity in
  reporting {BLEU} scores}.
\newblock In \emph{Proceedings of the Third Conference on Machine Translation:
  Research Papers}, pages 186--191, Brussels, Belgium. Association for
  Computational Linguistics.

\bibitem[{Rei et~al.(2020)Rei, Stewart, Farinha, and
  Lavie}]{rei-etal-2020-comet}
Ricardo Rei, Craig Stewart, Ana~C Farinha, and Alon Lavie. 2020.
\newblock \href {https://doi.org/10.18653/v1/2020.emnlp-main.213} {{COMET}: A
  neural framework for {MT} evaluation}.
\newblock In \emph{Proceedings of the 2020 Conference on Empirical Methods in
  Natural Language Processing (EMNLP)}, pages 2685--2702, Online. Association
  for Computational Linguistics.

\bibitem[{Sennrich et~al.(2016{\natexlab{a}})Sennrich, Haddow, and
  Birch}]{sennrich-etal-2016-edinburgh}
Rico Sennrich, Barry Haddow, and Alexandra Birch. 2016{\natexlab{a}}.
\newblock \href {https://doi.org/10.18653/v1/W16-2323} {{E}dinburgh neural
  machine translation systems for {WMT} 16}.
\newblock In \emph{Proceedings of the First Conference on Machine Translation:
  Volume 2, Shared Task Papers}, pages 371--376, Berlin, Germany. Association
  for Computational Linguistics.

\bibitem[{Sennrich et~al.(2016{\natexlab{b}})Sennrich, Haddow, and
  Birch}]{sennrich-etal-2016-neural}
Rico Sennrich, Barry Haddow, and Alexandra Birch. 2016{\natexlab{b}}.
\newblock \href {https://doi.org/10.18653/v1/P16-1162} {Neural machine
  translation of rare words with subword units}.
\newblock In \emph{Proceedings of the 54th Annual Meeting of the Association
  for Computational Linguistics (Volume 1: Long Papers)}, pages 1715--1725,
  Berlin, Germany. Association for Computational Linguistics.

\bibitem[{Stojanovski et~al.(2020)Stojanovski, Krojer, Peskov, and
  Fraser}]{stojanovski-etal-2020-contracat}
Dario Stojanovski, Benno Krojer, Denis Peskov, and Alexander Fraser. 2020.
\newblock \href {https://www.aclweb.org/anthology/2020.coling-main.417}
  {{C}ontra{CAT}: Contrastive coreference analytical templates for machine
  translation}.
\newblock In \emph{Proceedings of the 28th International Conference on
  Computational Linguistics}, pages 4732--4749, Barcelona, Spain (Online).
  International Committee on Computational Linguistics.

\bibitem[{Sun et~al.(2020)Sun, Wang, Zhou, Zhao, Huang, Chen, and
  Li}]{sun2020document}
Zewei Sun, Mingxuan Wang, Hao Zhou, Chengqi Zhao, Shujian Huang, Jiajun Chen,
  and Lei Li. 2020.
\newblock Capturing longer context for document-level neural machine
  translation: A multi-resolutional approach.
\newblock \emph{arXiv}, abs/2010.08961.

\bibitem[{Tiedemann and Scherrer(2017)}]{tiedemann-scherrer-2017-neural}
J{\"o}rg Tiedemann and Yves Scherrer. 2017.
\newblock \href {https://doi.org/10.18653/v1/W17-4811} {Neural machine
  translation with extended context}.
\newblock In \emph{Proceedings of the Third Workshop on Discourse in Machine
  Translation}, pages 82--92, Copenhagen, Denmark. Association for
  Computational Linguistics.

\bibitem[{Toral et~al.(2018)Toral, Castilho, Hu, and
  Way}]{toral-etal-2018-attaining}
Antonio Toral, Sheila Castilho, Ke~Hu, and Andy Way. 2018.
\newblock \href {https://doi.org/10.18653/v1/W18-6312} {Attaining the
  unattainable? reassessing claims of human parity in neural machine
  translation}.
\newblock In \emph{Proceedings of the Third Conference on Machine Translation:
  Research Papers}, pages 113--123, Brussels, Belgium. Association for
  Computational Linguistics.

\bibitem[{Tu et~al.(2018{\natexlab{a}})Tu, Liu, Shi, and
  Zhang}]{tu-etal-2018-learning}
Zhaopeng Tu, Yang Liu, Shuming Shi, and Tong Zhang. 2018{\natexlab{a}}.
\newblock \href {https://doi.org/10.1162/tacl_a_00029} {Learning to remember
  translation history with a continuous cache}.
\newblock \emph{Transactions of the Association for Computational Linguistics},
  6:407--420.

\bibitem[{Tu et~al.(2018{\natexlab{b}})Tu, Liu, Shi, and
  Zhang}]{Tu2018LearningTR}
Zhaopeng Tu, Yang Liu, Shuming Shi, and Tong Zhang. 2018{\natexlab{b}}.
\newblock \href {https://doi.org/10.1162/tacl_a_00029} {Learning to remember
  translation history with a continuous cache}.
\newblock \emph{Transactions of the Association for Computational Linguistics},
  6:407--420.

\bibitem[{Vaswani et~al.(2017)Vaswani, Shazeer, Parmar, Uszkoreit, Jones,
  Gomez, Kaiser, and Polosukhin}]{46201}
Ashish Vaswani, Noam Shazeer, Niki Parmar, Jakob Uszkoreit, Llion Jones,
  Aidan~N. Gomez, Lukasz Kaiser, and Illia Polosukhin. 2017.
\newblock \href
  {https://proceedings.neurips.cc/paper/2017/hash/3f5ee243547dee91fbd053c1c4a845aa-Abstract.html}
  {Attention is all you need}.
\newblock In \emph{Advances in Neural Information Processing Systems 30: Annual
  Conference on Neural Information Processing Systems 2017, December 4-9, 2017,
  Long Beach, CA, {USA}}, pages 5998--6008.

\bibitem[{Voita et~al.(2019)Voita, Sennrich, and Titov}]{voita-etal-2019-good}
Elena Voita, Rico Sennrich, and Ivan Titov. 2019.
\newblock \href {https://doi.org/10.18653/v1/P19-1116} {When a good translation
  is wrong in context: Context-aware machine translation improves on deixis,
  ellipsis, and lexical cohesion}.
\newblock In \emph{Proceedings of the 57th Annual Meeting of the Association
  for Computational Linguistics}, pages 1198--1212, Florence, Italy.
  Association for Computational Linguistics.

\bibitem[{Wang et~al.(2017)Wang, Tu, Way, and
  Liu}]{wang-etal-2017-exploiting-cross}
Longyue Wang, Zhaopeng Tu, Andy Way, and Qun Liu. 2017.
\newblock \href {https://doi.org/10.18653/v1/D17-1301} {Exploiting
  cross-sentence context for neural machine translation}.
\newblock In \emph{Proceedings of the 2017 Conference on Empirical Methods in
  Natural Language Processing}, pages 2826--2831, Copenhagen, Denmark.
  Association for Computational Linguistics.

\bibitem[{Zhang et~al.(2018)Zhang, Luan, Sun, Zhai, Xu, Zhang, and
  Liu}]{zhang-etal-2018-improving}
Jiacheng Zhang, Huanbo Luan, Maosong Sun, Feifei Zhai, Jingfang Xu, Min Zhang,
  and Yang Liu. 2018.
\newblock \href {https://doi.org/10.18653/v1/D18-1049} {Improving the
  transformer translation model with document-level context}.
\newblock In \emph{Proceedings of the 2018 Conference on Empirical Methods in
  Natural Language Processing}, pages 533--542, Brussels, Belgium. Association
  for Computational Linguistics.

\end{thebibliography}

\clearpage

\appendix

\section{Estimating CXMI}
\label{estimating-cxmi}

Let $S$ denote a random variable over source sentences, $T$ a random variable over target sentences and $C$ a random variable over possible context. We assume these random variables are distributed according to some \emph{true}, unknown distribution $p(\textbf{s}, \textbf{t}, \textbf{c})$. The cross-entropy between the true distribution $p$ and a probabilistic \emph{context-aware} neural translation model $q_{MT_C}(\textbf{t} | \textbf{s}, \textbf{c})$ is defined as
\begin{align*}
\text{H}_{q_{MT}}(T| S, C) &= \\
- \sum_{\textbf{s} \in V_S^*}&\sum_{\textbf{t} \in V_T^*}\sum_{\textbf{c} \in V_C^*}   p(\textbf{s}, \textbf{t}, \textbf{c}) \log 
q_{MT}(\textbf{s}, \textbf{t}, \textbf{c})  
\end{align*}
where $V_S^*, V_T^*, V_C^*$ represent the space of possible source sentences, target sentences and contexts respectively. Since we do not know the true distribution $p$, we cannot compute this quantity exactly. However, given a dataset of samples $\{(\textbf{s}^{(i)}, \textbf{t}^{(i)}, \textbf{c}^{(i)})\}^N_{i=0}$ assumed to be drawn from $p$, we can estimate this quantity using the Monte Carlo estimator
\begin{align*}
\text{H}_{q_{MT_C}}(T| S, C) &\approx \\
& - \frac{1}{N}\sum_{i=0}^N \log q_{MT_C}(\textbf{s}^{(i)}, \textbf{t}^{(i)}, \textbf{c}^{(i)})  
\end{align*}

If we consider the marginal $p(\textbf{s}, \textbf{t}) = \sum_{\textbf{c} \in V_C^*} p(\textbf{s}, \textbf{t}, \textbf{c})$, we can by a similar argument obtain an estimate for the cross-entropy for a \emph{context-agnostic} neural translation model $q_{MT_A}$ as:
\begin{align*}
\text{H}_{q_{MT_A}}(T| S) &\approx -\frac{1}{N} \sum_{i=0}^N \log q_{MT_A}(\textbf{s}^{(i)}, \textbf{t}^{(i)})  
\end{align*}

This leads trivially to the estimator for the cross-mutual information:
\begin{align*}
\text{\textsc{CXMI}}(C\rightarrow Y|X) & \approx \\
& \hspace{-1.5em} -\frac{1}{N} \sum_{i=1}^N \log \frac{q_{MT_A}(y^{(i)}|x^{(i)})}{q_{MT_C}(y^{(i)}|x^{(i)},C^{(i)})}
\end{align*}

\section{\textsc{CXMI} for $\text{EN}\rightarrow \text{FR}$}
\label{appendix:cxmi-enfr}

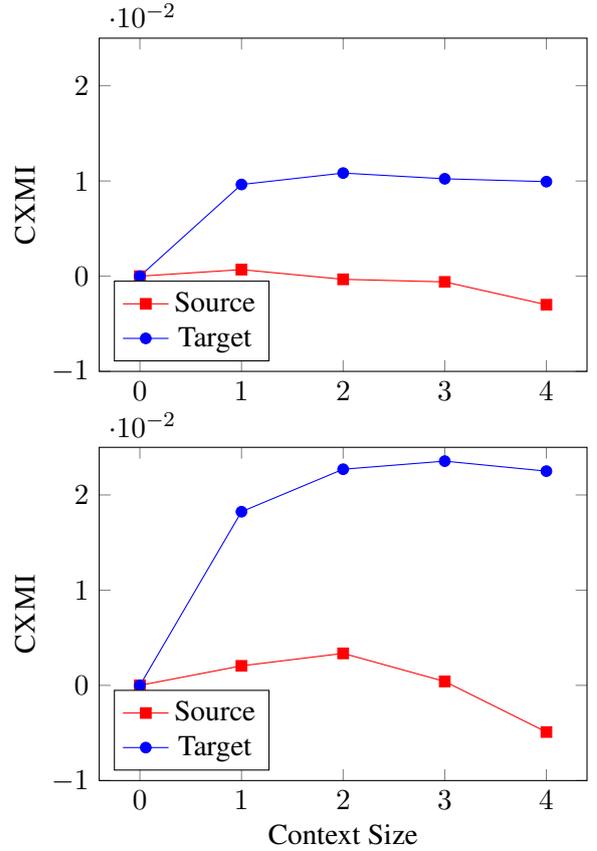
\begin{figure}[H]
\centering
\begin{tikzpicture}
\begin{axis}[
	x label style={at={(axis description cs:0.5,-0.1)},anchor=north},
    y label style={at={(axis description cs:-0.11,.5)},anchor=south},
	ylabel=\textsc{CXMI},
	xtick={0,1,2,3,4},
	ymin=-1e-2, ymax=2.5e-2,
	width=8cm,height=6cm,
    legend pos=south west
]
\addplot[color=red,mark=square*] coordinates {
    (0, 0)
	(1, 0.00069)
	(2, -0.00033)
	(3, -0.00060)
	(4, -0.00300)
};

\addplot[color=blue,mark=*] coordinates {
    (0, 0)
	(1, 0.00963)
	(2, 0.01083)
	(3, 0.01023)
	(4, 0.00993)
};
\legend{Source, Target}
\end{axis}
\end{tikzpicture}
\begin{tikzpicture}
\begin{axis}[
	xlabel=Context Size,
	ylabel=\textsc{CXMI},
	x label style={at={(axis description cs:0.5,-0.1)},anchor=north},
    y label style={at={(axis description cs:-0.11,.5)},anchor=south},
	xtick={0,1,2,3,4},
	ymin=-1e-2, ymax=2.5e-2,
	width=8cm,height=6cm,
    legend pos=south west
]
\addplot[color=red,mark=square*] coordinates {
    (0, 0)
	(1, 0.00205)
	(2, 0.00336)
	(3, 0.00041)
	(4, -0.0049)
};
\addplot[color=blue,mark=*] coordinates {
    (0, 0)
	(1, 0.01824)
	(2, 0.02270)
	(3, 0.02355)
	(4, 0.02250)
};
\legend{Source, Target}
\end{axis}
\end{tikzpicture}
\caption{$\text{\textsc{CXMI}}$ values as a function of source and target context sizes for non-pretrained (top) and pretrained (bottom) models for the $\text{EN}\rightarrow \text{FR}$ language pair.}
\label{fig:xmi-context-size-genera-en-frl}
\end{figure}

\section{Multi-Encoder}
\label{appendix:multiencoder}
For the multi-encoder model, we take the approach of initializing a separate transformer encoder for the context, with shared input-output embeddings with the original encoder (or decoder in the case of target context). The tokens in the current sentence attend to the context by the means of cross-attention. There are several other ways of formulation a multi-encoder context-aware systems, and exploring them is left for future research.
\end{document}